\documentclass[letterpaper, 10 pt, conference]{ieeeconf}  
\IEEEoverridecommandlockouts                              

\usepackage{graphicx}

\usepackage{stfloats}
\usepackage{newtxtext,newtxmath}
\usepackage[cal=cm]{mathalpha}
\usepackage{amsmath} 
\usepackage[hang]{subfigure} 
\usepackage{cite}
\usepackage{siunitx}
\usepackage{soul}
\usepackage{color}
\usepackage{xcolor}
\definecolor{softblack}{HTML}{333333}
\usepackage[ruled,linesnumbered]{algorithm2e}
\usepackage[font=scriptsize]{caption}

\soulregister\ref7  
\soulregister\cite7 
\soulregister\eqref7
\soulregister\prettyref7

\usepackage{multirow}
\usepackage{algpseudocode}
\usepackage{makecell}
\usepackage{booktabs}
\usepackage{threeparttable}
\usepackage{hyperref}
\usepackage{endnotes}

\newcommand{\eg}{\textit{e}.\textit{g}.~}

\bstctlcite{IEEEexample:BSTcontrol}

\usepackage{prettyref}
\newrefformat{fig}{Fig.~\ref{#1}}
\newrefformat{sec}{Sec.~\ref{#1}}
\newrefformat{tab}{Tab.~\ref{#1}}
\newrefformat{algo}{Algo.~\ref{#1}}
\newrefformat{eq}{(\ref{#1})}

\usepackage{color}

\makeatletter
\def\RAL@headerEven{IEEE ROBOTICS AND AUTOMATION LETTERS. PREPRINT VERSION. ACCEPTED JUNE, 2026}
\def\RAL@headerOdd{SUN \textit{et al.}: DPL: DEPTH-ONLY PERCEPTIVE HUMANOID LOCOMOTION}
\def\RAL@head{%
  \ifodd\value{page}\relax
    \ifnum\value{page}=1\relax
      \hbox{}\scriptsize\RAL@headerEven\hfil\thepage
    \else
      \hbox{}\scriptsize\RAL@headerOdd\hfil\thepage
    \fi
  \else
    \scriptsize\thepage\hfil\RAL@headerEven\hbox{}%
  \fi}
\def\ps@ralheadings{%
  \def\@oddhead{\RAL@head}\def\@evenhead{\RAL@head}%
  \def\@oddfoot{}\def\@evenfoot{}}
\makeatother

\title{\LARGE \bf
DPL: Depth-only Perceptive Humanoid Locomotion via \\ Realistic Depth Synthesis and Cross-Attention Terrain Reconstruction}

\vspace{-3mm}
\author{Jingkai Sun$^{1,2,\ast}$, Gang Han$^{1,\ast}$, Pihai Sun$^{1,3,\ast}$, Wen Zhao$^{1}$, \\ Jiahang Cao$^{2}$, Jiaxu Wang$^{4}$, Qiang Zhang$^{1,3,\dagger}$, Yijie Guo$^{1}$%
\thanks{Manuscript received: October 8, 2025; Revised: January 16, 2026; Accepted: June 9, 2026.}%
\thanks{This paper was recommended for publication by Editor Olivier Stasse upon evaluation of the Associate Editor and Reviewers comments.}%
\thanks{$^{1}$The authors are with Beijing Innovation Center of Humanoid Robotics Co. Ltd. {\tt\small Kale.Sun@x-humanoid.com}}%
\thanks{$^{2}$The authors are with The University of Hong Kong, Hong Kong, China.}%
\thanks{$^{3}$Artificial General Intelligence Institute, University of Science and Technology of China, Hefei, China.}%
\thanks{$^{4}$The author is with The Hong Kong University of Science and Technology, Hong Kong, China. $^{\ast}$ are equal contributors. $^{\dagger}$ is the corresponding author. {\tt\small jony.zhang@x-humanoid.com}}%
\thanks{Digital Object Identifier (DOI): see top of this page.}%
}
\vspace{-3mm}

\begin{document}
\maketitle

\thispagestyle{ralheadings}
\pagestyle{ralheadings}

\begin{abstract}
Recent advancements in legged robot perceptive locomotion have shown promising progress. However, terrain-aware humanoid locomotion remains largely constrained to two paradigms: depth image-based end-to-end learning and elevation map-based methods. The former suffers from limited training efficiency and a significant sim-to-real gap in depth perception, while the latter depends heavily on multiple vision sensors and localization systems, resulting in latency and reduced robustness. To overcome these challenges, we propose a novel framework that tightly integrates three key components: (1) Terrain-Aware Locomotion Policy with a Blind Backbone, which leverages pre-trained elevation map-based perception to guide reinforcement learning with minimal visual input; (2) Multi-Modality Cross-Attention Transformer, which reconstructs structured terrain representations from noisy depth images; (3) Realistic Depth Images Synthetic Method, which employs self-occlusion-aware ray casting and noise-aware modeling to synthesize realistic depth observations, achieving over 30\% reduction in terrain reconstruction error. This combination enables efficient policy training with limited data and hardware resources, while preserving critical terrain features essential for generalization. We validate our framework on a full-sized humanoid robot, demonstrating agile and adaptive locomotion across diverse and challenging terrains. 

\end{abstract}


\vspace{-3mm}
\section{Introduction}

\PARstart{H}{umanoid} robots offer immense potential for enabling autonomous mobility in human-centric, unstructured environments. Achieving this vision requires the development of perceptive locomotion systems that integrate visual perception and control, enabling real-time gait adaptation to complex terrain. While recent progress has shown that quadrupedal robots can robustly traverse cluttered, uneven, and deformable surfaces, transferring these capabilities to humanoid platforms introduces a fundamentally different set of challenges.

Current solutions for terrain-aware humanoid locomotion typically fall into two paradigms. The first directly maps depth images to control actions via end-to-end learning. Although elegant in design, such methods are often hindered by sim-to-real gaps and poor generalization, especially under sensor noise and occlusion~\cite{zhuang2023robot, zhuang2025humanoid, yu2024walking, luo2024pie, rudin2025parkour, duan2024learning}. The second class of methods reconstructs elevation maps by traditional methods and performs planning or control in the structured geometry space~\cite{he2025attention, allshire2025visual, sun2025learning, long2025learning, wang2025beamdojo}. While effective in simulation, these approaches require multiple exteroceptive sensors and accurate localization, rendering them susceptible to delays and drift in real-world deployments. In addition to the aforementioned limitations, elevation map-based methods often struggle with occluded or blind regions~(\eg gaps). To address these issues, hand-crafted rules are typically introduced to infer traversability in such regions. However, these manual heuristics increase system complexity and may introduce conflicts or inconsistencies between rules. Our proposed approach is designed to address all of these challenges in a unified and learning-based manner.

To address these challenges, we present a unified framework for perceptive humanoid locomotion that combines structured terrain reasoning with end-to-end reinforcement learning. Compared with the prior terrain reconstruction-based bipedal locomotion method~\cite{duan2024learning}, our proposed approach is designed to operate under noisy, partial, and self-occluded depth perception, without relying on global localization. The framework integrates three key components. First, a \textit{terrain-aware locomotion policy with blind backbone} leverages pre-trained elevation-based priors to guide locomotion learning without requiring vision input at runtime. Second, a \textit{multi-modal cross-attention transformer} reconstructs local terrain geometry from noisy first-person depth images and proprioceptive state history. Third, a \textit{realistic depth images synthetic method} synthesizes realistic sensor observations using ray casting with self-occlusion-aware geometry and noise modeling, significantly narrowing the domain gap between simulation and reality. Unlike prior work~\cite{duan2024learning}, which utilizes synthetic depth primarily for offline data collection to train a decoupled reconstruction module, our approach fully integrates the depth synthesis pipeline into the reinforcement learning loop. This design enables end-to-end fine-tuning where the locomotion policy interacts directly with the reconstructed results rather than grounded height maps, thereby allowing the agent to learn robust behaviors that explicitly adapt to perception-induced errors and latency during the training phase.

We evaluate the proposed framework on a full-sized humanoid robot across a variety of challenging scenarios, including slopes, stairs, gaps, and uneven outdoor surfaces. Our system demonstrates robust and agile locomotion in both simulation and the real world, even under strong perception degradation.

\noindent
The key contributions of this work are summarized as follows:
\begin{itemize}
    \item We propose a multi-stage training framework for depth-based humanoid perceptive locomotion that enables end-to-end fine-tuning based on pre-trained policies without relying on external localization systems.
    \item We introduce a cross-modal transformer that reconstructs terrain geometry from partial depth and proprioceptive inputs.
    \item We develop a realistic depth image synthetic method that simulates occlusion-aware and noise-corrupted depth images for efficient and realistic policy training.
    \item We validate our system on a real humanoid platform, demonstrating superior terrain generalization and robustness.
\end{itemize}
\vspace{-4mm}
\section{Related Work}
Perception plays a crucial role in enabling stable and precise locomotion. Unlike blind control methods, perceptive locomotion empowers robots to anticipate and respond to terrain features before physical contact occurs. Several works~\cite{imai2022vision,yang2021learning} have adopted depth-image-based frameworks for quadrupedal locomotion, using Transformer architectures or other fusion networks to integrate perceptual and proprioceptive inputs. Compared to quadruped robots, training humanoid robots typically requires more data and longer durations. This makes training from scratch highly demanding in terms of computational resources and time, especially when depth images are obtained inefficiently. Our proposed pretraining-based approach significantly improves training efficiency. Other approaches~\cite{zhuang2023robot,cheng2024extreme} convert depth images into height maps through a two-stage distillation process, applying extensive data augmentation to reduce the sim-to-real gap. However, these methods are often constrained by limited sample efficiency and a persistent domain gap between synthetic and real-world depth images. Our proposed method aims to improve the training efficiency of depth-based and bridge the sim-to-real gap. \cite{duan2024learning} combines a blind policy and a vision-based modulator. This approach utilizes the combined policy to collect depth data for training the reconstruction module. The reconstructed terrain still suffers from a domain gap between the vision-based modulator, which is not finetuned jointly. Our depth generation methods support the end-to-end finetuning to improve performance. Some elevation map-based approaches utilize depth sensors and localization systems to construct terrain maps and identify salient regions for locomotion~\cite{he2025attention, allshire2025visual, sun2025learning, zhang2025distillation, hoeller2024anymal,hoeller2022neural,chen2025learning}. These methods depend heavily on multi-sensor fusion or accurate pose estimation, introducing complexity and limiting robustness. Moreover, the reliance on external localization systems constrains their update frequency, making them less suitable for dynamic scenarios. In contrast, our method requires only a single depth camera and achieves significantly higher update frequency, enabling more responsive and efficient control.

\section{Preliminary}
We formulate humanoid locomotion as a Markov Decision Process (MDP) defined by the tuple $(\mathcal{S}, \mathcal{A}, \mathcal{R}, p, \gamma)$. Here, $\mathcal{S}$ denotes the state space, $\mathcal{A}$ the action space, and $\mathcal{R}$ the reward function. The system dynamics are captured by the transition probability $p(s_{t+1} | s_t, a_t)$, which governs the evolution from the current state $s_t$ to the next state $s_{t+1}$ after executing action $a_t$. The discount factor $\gamma \in [0,1]$ balances immediate and future rewards. At each time step $t$, the policy $\pi_\theta(a_t|s_t)$ selects an action based on the current state, producing a trajectory $\tau = (s_0,a_0,\dots,s_{T-1},a_{T-1})$. The objective is to optimize the policy parameters $\theta$ to maximize the expected cumulative discounted reward:

\begin{equation}
\arg\max_{\theta} \mathbb{E}{(s_t,a_t)\sim p\theta} \left[ \sum_{t=0}^{T-1} \gamma^t r_t \right].
\end{equation}

To enable more natural and human-like interactions between the robot and its environment, we incorporate the Adversarial Motion Prior (AMP) framework~\cite{peng2021amp}. Unlike methods that directly track reference joint trajectories, AMP encourages the policy to generate actions in a human-consistent style. This is achieved by introducing a discriminator \( D(s_t, a_t) \) that distinguishes between state transitions from reference demonstrations and those generated by the policy.

The reward from the discriminator is defined as:
\begin{equation}
    r_i = \max\left[0,\; 1 - \frac{1}{4} \left(D(s_t^I, s_{t+1}^I) - 1\right)^2 \right]
\end{equation}
where \( s_t^I \) denotes the partial state observations provided to the discriminator.

In addition to the standard reinforcement learning objectives, the AMP loss is formulated as:
\begin{equation}
\begin{aligned}
    L_i =\;& \frac{1}{2} \mathbb{E}_{(s_t^I, s_{t+1}^I) \sim \mathcal{D}} \left[\left(D(s_t^I, s_{t+1}^I) - 1\right)^2 \right] \\
    +\;& \frac{1}{2} \mathbb{E}_{(s_t^I, s_{t+1}^I) \sim \pi} \left[\left(D(s_t^I, s_{t+1}^I) + 1\right)^2 \right] \\
    +\;& \lambda_{\text{GP}} \mathbb{E}_{(s_t^I, s_{t+1}^I) \sim \mathcal{D}} \left[\left\| \nabla D(s_t^I, s_{t+1}^I) \right\|^2 \right]
\end{aligned}
\end{equation}

Here, \( (s_t^I, s_{t+1}^I) \sim \mathcal{D} \) and \( (s_t^I, s_{t+1}^I) \sim \pi \) represent state transitions sampled from the demonstration dataset and the policy, respectively. The final term imposes a gradient penalty weighted by \( \lambda_{\text{GP}} \), which stabilizes the adversarial training process. In our study, the motion priors are constructed using motion capture data from two widely recognized public repositories, the SFU Mocap dataset and the CMU Mocap dataset.





\vspace{-3mm}
\section{Method}
\begin{figure}[t]
    \centering
    \vspace{2mm}
    \includegraphics[width=0.99\linewidth]{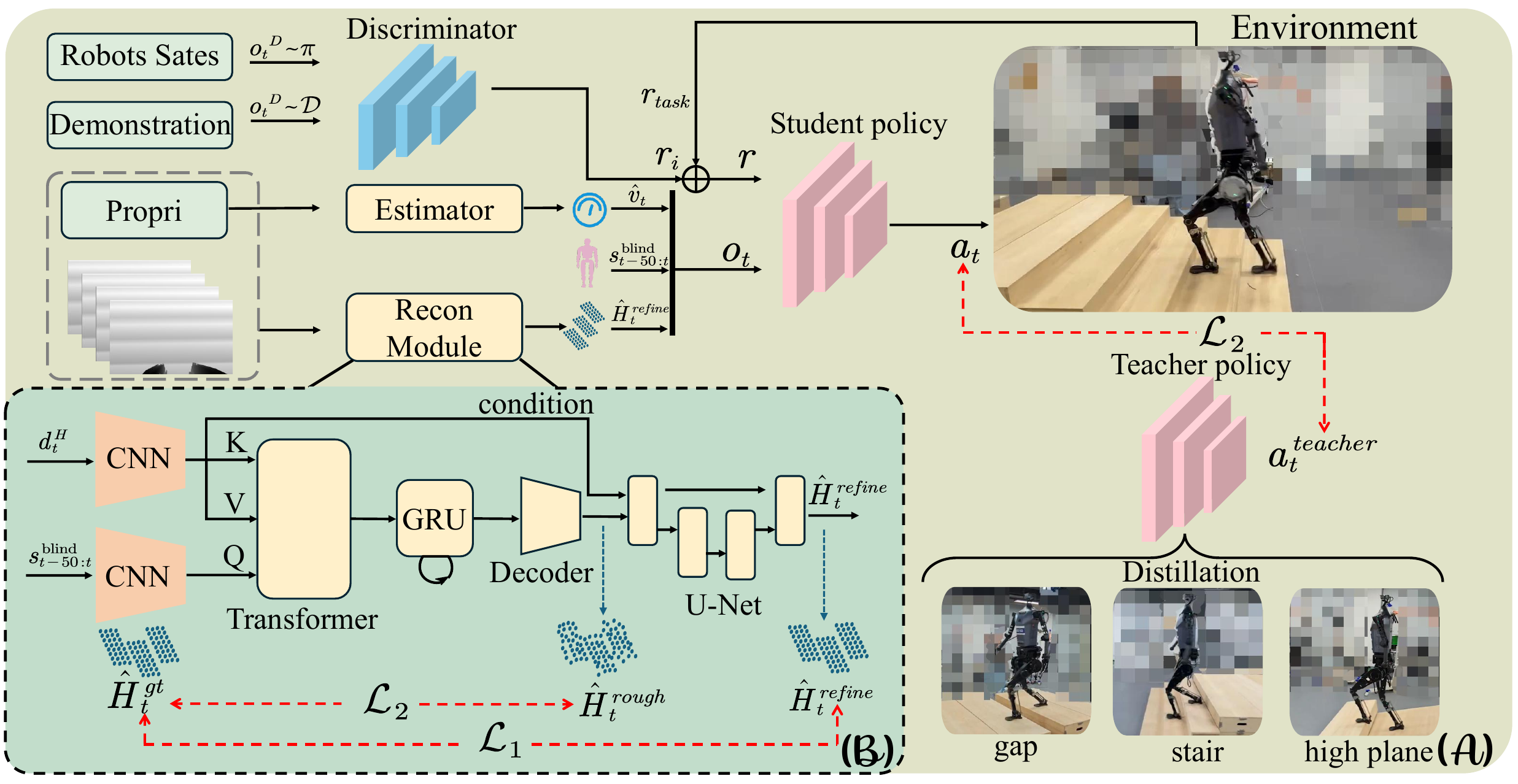}
\caption{\textbf{Overview of the proposed teacher–student distillation framework for humanoid perceptive locomotion.} \textbf{\textit{(A)}} The student policy interacts with the environment to generate actions, while a teacher policy provides supervision via an $\mathcal{L}_2$ loss. The distillation process transfers locomotion skills across diverse terrains. \textbf{\textit{(B)}} A reconstruction module integrates proprioceptive history and depth information through a Transformer–based pipeline to refine terrain representations $\hat{H}^{\text{refine}}_t$, supervised by ground truth. The refined features are fed into the student policy, which is further optimized using distillation and adversarial reinforcement learning.}
    \label{Overview}
\end{figure}

We propose a unified framework for humanoid perceptive locomotion that leverages teacher–student distillation to transfer robust skills across diverse terrains as shown in Fig.~\ref{Overview}. The system consists of a Transformer-based reconstruction module, a realistic depth noise model, and a locomotion policy. The pipeline begins with the reconstruction module, which predicts local heightmaps from egocentric depth inputs. To bridge the sim-to-real gap, we introduce a stochastic noise model that simulates sensor artifacts. Finally, the locomotion policy is trained to navigate using these reconstructed, noise-augmented representations, ensuring robustness against perceptual uncertainty and artifacts.

\subsection{Multi-Modality Cross Attention Transformer for Terrain Reconstruction}
The module receives proprioceptive history $s^{\text{perc}}_{t-50:t}$ and temporal depth observations $d^{H}_t$ ($H=5$) as inputs, and predicts the surrounding heightmap in the robot’s local frame.

As shown in Fig.~\ref{Overview}~(B), depth images are first compressed through a convolutional encoder to obtain compact spatial features, while the proprioceptive history is embedded into latent vectors that capture the robot’s kinematic and dynamic states. These two modalities are fused through a cross-attention transformer, in which proprioceptive embeddings serve as the query and depth embeddings provide keys and values:
\begin{equation}
z^\text{fused}_t = \operatorname{Attn}\!\left(Q = z^\text{prop}_t,\; K = z^\text{depth}_t,\; V = z^\text{depth}_t\right).
\end{equation}

This operation enables the reconstructor to selectively emphasize terrain features most relevant to the robot’s current motion state. Intuitively, the depth map alone provides only a partial and noisy view of the terrain, while proprioceptive inputs encode gait phase, body posture, and velocity. By aligning proprioceptive queries with contextual depth features, the model selectively emphasizes critical regions in the visual stream for reconstruction, on the robot states. The fused features are subsequently passed through recurrent memory units to maintain temporal consistency and accumulate knowledge of unobserved terrain. A decoder then produces a rough heightmap $\hat{H}^\text{rough}_t$, which is supervised by mean squared error against the ground-truth local heightmap $H^{gt}_t$. However, as shown in prior work \cite{duan2024learning}, rough reconstructions often suffer from blurred edges and non-flat surfaces, which limit their utility as policy inputs. To address this, we employ a conditional U-Net that takes both the rough prediction and encoded depth latent as input and outputs a refined reconstruction:
\begin{equation}
\hat{H}^\text{refined}_t = \text{U-Net}(\hat{H}^\text{rough}_t,z^\text{depth}_t).
\end{equation}
The refined heightmap is optimized with L1 loss to sharpen edges and improve flatness, particularly in regions of terrain transitions. The overall loss thus combines both stages:
\begin{equation}
\mathcal{L} = | \hat{H}^\text{rough}_t - H^{gt}_t |_2^2 + | \hat{H}^\text{refined}_t - H^{gt}_t |_1.
 \end{equation}
Through this design, the terrain reconstructor not only improves the fidelity of heightmap predictions but also enforces the vision encoder to learn terrain-sensitive representations. 
\subsection{Realistic Depth Images Synthetic Method}
\subsubsection{Synthetic Depth via GPU Ray Casting}
We synthesize depth images by tracing virtual rays from calibrated pinhole cameras into a dynamic 3D triangle mesh composed of a static terrain and articulated robot geometry. Each image is generated at a resolution of $600 \times 480$ across $B$ parallel environments.

The terrain mesh, denoted $\mathcal{M}^\text{ter}$, is shared by all environments and may be either procedurally generated or loaded from an external mesh or heightfield. To ensure consistent alignment with the global coordinate frame used by downstream modules(specifically within the Isaac Gym~\cite{makoviychuk2021isaac} terrain generation protocol), the terrain vertices are translated by a fixed offset $\mathbf{o}^\text{ter} = [-b, -b, 0]^\top$, where $b$ denotes the border margin:
\begin{equation}
\mathbf{v}_i^\text{world} \leftarrow \mathbf{v}_i^\text{ter} + \mathbf{o}^\text{ter}.
\end{equation}
Robot geometry is constructed by parsing its kinematic tree and aggregating all visual meshes at the rigid-body level. For each visual element, the local mesh transform is composed with the accumulated joint transforms to express all vertices in their corresponding body frames. This yields a canonical mesh template,

\begin{equation} \begin{aligned} &\{\mathbf{V}^\text{loc},\ \mathbf{F},\ \mathbf{b}\}, \\ &\mathbf{V}^\text{loc}\!\in\!\mathbb{R}^{N_v\times3}, \\ &\mathbf{F}\!\in\!\mathbb{N}^{N_f\times3}, \\ &\mathbf{b}\!\in\!\{0,\dots,N_\text{body}{-}1\}^{N_v}, \end{aligned} \end{equation}
where $\mathbf{V}^\text{loc}$ stores per-vertex coordinates in the local body frame, $\mathbf{F}$ denotes the face connectivity, and $\mathbf{b}$ maps each vertex to its associated rigid body. At each simulation step $t$, the rigid-body states for environment $e$ provide body-frame poses $(\mathbf{q}^{(e)}_j, \mathbf{t}^{(e)}_j)$ for all $j=0,\dots,N_{\text{body}{-}1}$, where $\mathbf{q}_j \in \mathbb{H}$ is the orientation quaternion and $\mathbf{t}_j \in \mathbb{R}^3$ is the translation. Vertex positions in world coordinates are obtained via rigid-body transformations:
\vspace{-2mm}
\begin{equation}
\mathbf{v}^{(e)}_k(t)
=
\mathbf{R}\left(\mathbf{q}^{(e)}_{b_k}\right)\, \mathbf{v}^\text{loc}_k
+
\mathbf{t}^{(e)}_{b_k},
\end{equation}

where $\mathbf{R}(\cdot)$ denotes the rotation matrix corresponding to a unit quaternion. These transformed vertices define the robot mesh for each environment, which is combined with the static terrain for ray-traced rendering.

Given camera intrinsics $(f_x, f_y, c_x, c_y)$ and extrinsics $(\mathbf{R}_{cw}, \mathbf{t}_{cw})$ mapping world coordinates to the camera frame, we cast one ray per pixel $(u, v)$ on the $600 \times 480$ grid. The direction in normalized camera coordinates is given by:
\begin{equation}
x = \frac{u - c_x}{f_x},\quad y = \frac{v - c_y}{f_y},\quad
\mathbf{d}_c = \frac{[x,y,1]^\top}{|[x,y,1]|},
\end{equation}

and the corresponding origin and direction in world coordinates are:
\vspace{-2mm}
\begin{equation}
\mathbf{o}_w = \mathbf{R}_{cw}^\top (-\mathbf{t}_{cw}),\quad
\mathbf{d}_w = \mathbf{R}_{cw}^\top \mathbf{d}_c.
\end{equation}
These rays are intersected with the union of the terrain and robot mesh to determine the first valid surface hit. The intersection returns the smallest positive distance $t^\star$ along $\mathbf{d}_w$, or a no-hit signal. The 3D hit location in world frame is
\begin{equation}
\mathbf{p}_w = \mathbf{o}_w + t^\star \mathbf{d}_w
\end{equation}
and the corresponding depth value is computed as the $z$-component in camera frame:
\begin{equation}
z(u,v) = \mathbf{e}_z^\top \left( \mathbf{R}_{cw} \mathbf{p}_w + \mathbf{t}_{cw} \right).
\end{equation}
This geometric depth synthesis module plays a foundational role in our perception pipeline by providing accurate and physically consistent depth observations for all simulated environments. 
\subsubsection{Depth Domain Randomization with Noise Model}

To bridge the gap between idealized ray-traced depth and realistic sensor outputs, we introduce a stochastic corruption model that synthesizes key artifacts observed in real-world depth images. The model accounts for both continuous noise and structured missingness. 


To mitigate border distortions and invalid rays near the image perimeter, we crop a fixed margin of $M$ pixels from all sides and resample the central region to the target resolution using a differentiable interpolation. Out-of-range values outside the sensor’s valid interval $[z_{\min}, z_{\max}]$ are clipped prior to noise injection, in accordance with empirical calibration studies~\cite{nguyen2012modeling}. Following empirical studies on Kinect, the depth variance increases with range and grows sharply near grazing angles. We adopt the parametric family in~\cite{nguyen2012modeling} and re-center by the frame mean $\mu_z$ for scene adaptivity rather than a fixed value:
\vspace{-1mm}
\begin{equation}
\sigma_z(z,\theta)\;=\;a\;+\;b\,(z-\mu_z)^2\;+\;\frac{c}{\sqrt{z}},    
\end{equation}
with $a,b,c$ are hyperparameters. Axial noise is injected additively,
\vspace{-1mm}
\begin{equation}
\hat{D}\;=\;\tilde{D}\;+\;\mathcal{N}\big(0,\sigma_z^2\big), 
\end{equation}
mirroring the quadratic-with-range behavior reported in controlled measurements and exploited for KinectFusion weighting~\cite{nguyen2012modeling, khoshelham2012accuracy}. Lateral uncertainty is approximately range-independent in pixel units but scales linearly in meters after reprojection. We approximate this as a small, range-proportional perturbation:
\begin{equation}
\sigma_L(z)\;\approx\;\alpha\,z\,\xi,\quad \xi\sim\mathcal{U}[-1,1], 
\end{equation}
which captures mixed-pixel edge fattening and spatial smearing in metric space without explicitly resampling neighbors. This matches empirical characterizations contrasting pixel-space versus metric-space growth~\cite{khoshelham2012accuracy
,mallick2014characterizations}.

Depth holes appear more frequently where measurement variance is high. We form a per-pixel uncertainty proxy by combining axial and lateral components

\begin{equation}
\sigma_{\text{tot}}\,=\,\sqrt{\,w\,\sigma_z^2 + \sigma_L^2\,},  \end{equation}

To reflect the spatially varying reliability of depth measurements, we construct a per-pixel uncertainty map $\sigma_{\text{tot}}$ that aggregates both axial and lateral components, with axial error amplified by a weighting factor $w \ge 1$ to emphasize its disproportionate influence in volumetric reconstruction tasks. To ensure adaptivity across different scenes, we normalize this map by the frame-wise maximum uncertainty before scaling it to a pixel-wise dropout probability $p_\sigma \in [0, \rho]$, where $\rho$ denotes the maximum missing-data ratio. Pixels are then stochastically invalidated by comparing $p_\sigma$ against uniformly distributed random noise, simulating realistic uncertainty-induced corruption patterns.

Crucially, invalid depth measurements predominantly cluster along geometric discontinuities, where grazing incidence angles and mixed-pixel effects cause severe signal degradation. To accurately emulate this structural noise, we first compute the spatial gradient of the depth map using $3\times3$ Sobel operators. Let $G_x$ and $G_y$ denote the gradient components obtained via convolution in the horizontal and vertical directions, respectively. The raw gradient magnitude $G_{\text{raw}} = \sqrt{G_x^2 + G_y^2}$ is subsequently convolved with a $3\times3$ mean kernel filter to produce a smoothed gradient map $G$. This pre-filtering step is essential for spatially aggregating gradient information and suppressing high-frequency texture noise that does not correspond to physical geometry. To identify salient edges robustly across varying scene scales, we employ a frame-adaptive quantile thresholding strategy. Specifically, we calculate the dynamic threshold $\tau_{0.8}$ corresponding to the 80-th percentile of gradient magnitudes within the current frame. This isolates the set of significant structural edges $\mathcal{E} = \{ (u,v) \mid G(u,v) > \tau_{0.8} \}$, ensuring that dropout targets only the top 20\% of discontinuities regardless of the absolute depth range. Within this edge set, we define a dropout probability map $p_e$ by normalizing the gradient magnitude relative to the local maximum:\begin{equation}p_e(u,v) = \lambda_e \cdot \frac{G(u,v)}{\max_{(i,j) \in \mathcal{E}} G(i,j) + \epsilon} \cdot \mathbf{1}{(u,v) \in \mathcal{E}},\end{equation}where $\lambda_e$ governs the peak dropout intensity. This formulation ensures that sharper edges (higher gradients) are assigned proportionally higher probabilities of invalidation. Finally, the edge-based probability is combined with the uncertainty-based probability $p_\sigma$, and the binary validity mask is generated via a stochastic Bernoulli trial: $\mathcal{M} = \mathbf{1}\{\text{rand} < \text{clamp}(p_\sigma + p_e, 0, 1)\}$. The final output $D_{\text{out}} = \hat{D} \odot (1 - \mathcal{M})$ thus exhibits realistic, structurally correlated holes along object boundaries.



\subsection{Terrain-Aware Locomotion Policy with Blind Backbone}
\noindent \textbf{Observation and Action Space.}
In humanoid locomotion tasks, the observation vector integrates proprioceptive feedback, environmental perception, user-specified commands, and periodic signals to provide a comprehensive state description. Proprioceptive information includes its angular velocity $\omega_t$, and the position and velocity of all measurable actuated joints $q_t, \dot{q_t}$, together with the projected gravity $g_t$ and linear velocity $v_t$ obtained by estimator, following~\cite{zhang2025distillation}. To encode terrain structure, a heightmap $\textbf{h}_t$ covering a $1.0\text{m} \times 1.0\text{m}$ region in front of the robot at $5\text{cm}$ resolution is provided, representing relative elevations with respect to the floating base. This encoding adapts naturally to vertical oscillations during gait cycles and eliminates the need for global mapping or odometry. The command input $\textbf{c}_t = (v^x_t, v^y_t, \omega^{\text{yaw}}_t)$ specifies the target linear and yaw velocities. Periodic signals are added to structure gait generation: for each leg, sine and cosine functions of a phase variable $\phi_t$ are shifted by a leg-specific offset $\gamma^i_t$ ($i \in {\text{left}, \text{right}}$). Finally, the previous action is appended to the observation to promote temporal smoothness. The policy maps these observations to actions consisting of the desired joint targets and delta command $a_t \in \mathbb{R}^{22}$, updated at $100\text{Hz}$ and tracked by low-level PD controllers at $1\text{kHz}$.
In addition to joint actions, the policy adapts locomotion by adjusting the desired forward velocity and the gait phase.

\noindent \textbf{Policy Structure.}
We adopt a teacher-student architecture shown in Fig. \ref{Overview}~(A) that combines a pretrained blind policy and a vision-based modulator, both in the student and teacher policy. The blind policy provides a stable baseline locomotion controller, while the vision-based modulator adapts the baseline to complex environments through terrain perception. The blind policy $\pi_{\text{blind}}$ takes proprioceptive states and user commands as input, excluding the terrain heightmap:
\begin{equation}
s^{\text{blind}}_t = \big[ q_t, \dot{q}_t, \omega_t, g_t, c_t,\phi_t,\hat{v}_t \big],
\end{equation}
and outputs joint actions
\begin{equation}
a^{\text{blind}}_t = \pi_{\text{blind}}(s^{\text{blind}}_t), \quad a^{\text{blind}}_t \in \mathbb{R}^{20}.
\end{equation}
The perceptive policy $\pi_{\text{perc}}$ augments the blind backbone by additionally incorporating the terrain heightmap and the blind action:
\begin{equation}
s^{\text{perc}}_t = \big[ s^{\text{blind}}_t,  h_t, a^{\text{blind}}_t \big].
\end{equation}
It outputs a modulating joint action $a^{\text{mod}}_t \in \mathbb{R}^{20}$, and residual actions that adjust the locomotion clock and velocity commands. The final joint action $a^j_t$ is a convex combination of blind and modulated actions:
\begin{equation}
a^j_t = (1-\alpha) a^{\text{mod}}_t + \alpha a^{\text{blind}}_t, \quad \alpha \in [0,1].
\end{equation}
For gait phase control, the perceptive policy outputs a residual increment $\delta \phi_t$:
\begin{equation}
\Delta \phi_t = \text{clip}\big(\delta \phi_t, \Delta \phi_{\min}, \Delta \phi_{\max}\big) + \Delta \phi^{\text{cmd}},
\end{equation}
and for command adaptation, it also outputs a residual velocity $\delta v^x_t$:
\begin{equation}
v^{x, \text{mod}}_t = \text{clip}\big(\delta v^x_t, v^x_{\min},  v^x_{\max}\big) + v^x_t.
\end{equation}
This structure enables the perceptive policy to refine the blind baseline by adjusting both motor targets and high-level gait variables, ensuring robustness in simple settings while adapting effectively to irregular terrains.

Our reward functions encourage the humanoid robot to follow commanded velocities, maintain stable orientation, and achieve smooth and efficient locomotion. In addition to classical terms adapted from prior gait phase-based reward~\cite{zhang2024whole}, we design several new rewards specifically for humanoid gaits. The stumble and stumble during swing terms penalize undesired impacts of feet with obstacles or terrain edges.
All detailed task reward functions $\mathcal{T}$ used in our training framework are summarized in Table~\ref{tab:reward}. The rewards only applied in vision stage are highlighted with \textcolor{cyan}{blue} color.

We adopt a teacher-student distillation paradigm to train a general-purpose locomotion policy capable of robustly traversing diverse terrains. Terrain-specific expert policies, each specialized for a particular environment such as gaps, stairs, or flat planes, are first trained with privileged information and full-state observability. During student policy training, these expert actions are used as supervision via $\mathcal{L}_2$ loss, guiding the student to imitate the experts under partial observation. We refer to this training paradigm, which leverages multiple expert policies for supervision, as the "Multi-teacher" framework. Crucially, unlike the experts, the student operates on noisy, reconstructed heightmaps generated by our vision module. By jointly optimizing the PPO~\cite{schulman2017proximal} objective alongside the distillation loss, the student goes beyond merely cloning the experts and explicitly fine-tunes its behavior to adapt to perception-induced uncertainties such as latency and geometric distortions. This fine-tuning process is essential for closing the sim-to-real gap, enabling the policy to map imperfect observations directly to robust actions.
\begin{table}[t]
\centering
\vspace{2mm}
\caption{Humanoid Reward Terms}
\scalebox{1.00}{
\begin{tabular}{l l}
\toprule
\textbf{Reward} & \textbf{Equation ($r_i$)} \\
\midrule
X velocity diff & $\exp(-3|v^{cmd}_x - \bar{v}_x|)$ \\
Y velocity diff & $\exp(-10|v^{cmd}_y - \bar{v}_y|)$ \\
Z velocity diff & $\exp(-2|v_z|)$ \\
Angular velocity & $\exp(-\|\omega_{xy}\|^2)$ \\
Orientation diff & $\exp(-100\|g_{proj,xy}\|^2)$ \\
Torques penalty & $\sum (\tau/k_P)^2$ \\
Joint velocity penalty & $\sum \dot{q}^2$ \\
DoF pos limits & $\sum \text{clip}(q-q_{lim})$ \\
Torque limits & $\exp(-0.005\|\tau-\tau_{max}\|)$ \\

\textcolor{cyan}{Delta v command limits } &\textcolor{cyan}{ $\exp(-200|a_{22}|)$} \\
\textcolor{cyan}{Delta cycle limits } &\textcolor{cyan}{ $\exp(-200|a_{21}|)$ } \\
\textcolor{cyan}{Delta command smoothness } &\textcolor{cyan}{ $\|\Delta a_{[21,22]}\|$ } \\
\textcolor{cyan}{Stumble } &\textcolor{cyan}{ $\text{IF}{\{|F^x_{foot}|>0.5F^z_{foot}\}}$  }\\
\textcolor{cyan}{Stumble during swing } &\textcolor{cyan}{ $\text{IF}{\{|F^x_{foot}|>10\}}$  }\\

\bottomrule
\end{tabular}
}
\label{tab:reward}
\end{table}

\section{Experiments}
For our experiments, we employ the full-sized humanoid robot \textit{TienKung Ultra}, which features 20 actuated degrees of freedom (DOF), including active joints in the legs and arms. We employ an Orbbec 335L depth camera as the vision sensor. 
The pretrained control policy is optimized using Proximal Policy Optimization~(PPO)~\cite{schulman2017proximal} and trained on a single NVIDIA RTX 4090 GPU with 4096 parallel instances. The terrain reconstructor module is trained on an A100 GPU with 2048 parallel environments in Isaac Gym~\cite{makoviychuk2021isaac}.
\subsection{Simulation Results}
\begin{table*}[t]
\centering
\caption{Mean Absolute Error (MAE) [cm] across terrain types}
\scalebox{1.0}{
\begin{tabular}{lccccccc}
\toprule
\textbf{Method} & \textbf{Rough Slope Down} & \textbf{Rough Slope Up} & \textbf{Stairs Down} & \textbf{Stairs Up} & \textbf{High Plane} & \textbf{Discrete} & \textbf{Gap} \\
\midrule
Ours & \textbf{2.88$\pm$0.15} & \textbf{2.29$\pm$0.17} & \textbf{3.66$\pm$0.43} & 4.51$\pm$0.51 & \textbf{3.24$\pm$0.27} & \textbf{3.18$\pm$0.28} & \textbf{4.21$\pm$0.45} \\
w/o GRU & 3.83$\pm$0.37 & 3.17$\pm$0.32 & 4.06$\pm$0.39 & 5.23$\pm$0.48 & 5.08$\pm$0.43 & 3.45$\pm$0.37 & 5.06$\pm$0.51 \\
w/o Condition & 4.32$\pm$0.42 & 5.14$\pm$0.52 & 5.84$\pm$0.60 & 6.02$\pm$0.63 & 6.23$\pm$0.78 & 6.29$\pm$0.80 & 5.72$\pm$0.62 \\
CNN-based~\cite{yu2024walking} & 3.01$\pm$0.22 & 2.83$\pm$0.19 & 4.33$\pm$0.42 & 4.64$\pm$0.43 & 4.07$\pm$0.34 & 5.03$\pm$0.49 & 4.73$\pm$0.47 \\
ResNet-based~\cite{duan2024learning} & 2.95$\pm$0.18 & 2.76$\pm$0.16 & 4.03$\pm$0.33 & \textbf{4.42$\pm$0.42} & 3.27$\pm$0.36 & 3.63$\pm$0.42 & 4.21$\pm$0.43 \\
\bottomrule
\end{tabular}
}
\vspace{-2mm}
\label{tab:mae_terrain}
\end{table*}

\begin{figure}[t]
    \centering
    \vspace{-2mm}
    \includegraphics[width=0.90\linewidth]{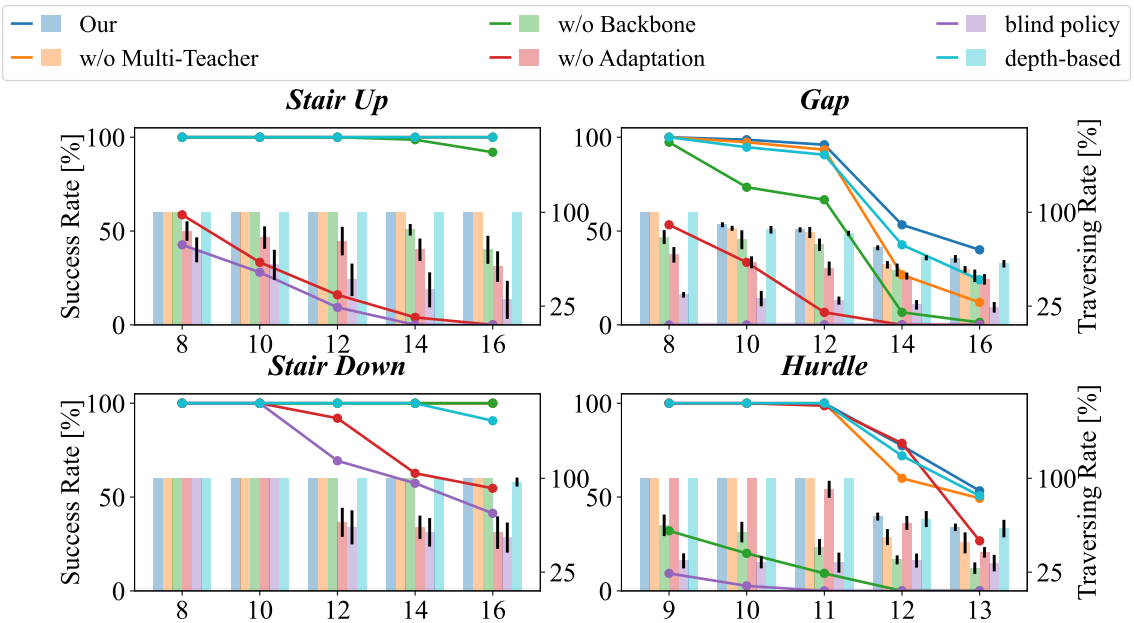}
\caption{Ablation study of the proposed framework across four challenging terrains: Stair Up, Gap, Stair Down, and Hurdle. All success rates and traversing rates are calculated over 100 independent environments for each terrain type and difficulty level. Success rate (lines) and traversing rate (bars) are reported for our full model and three ablated variants: without multi-teacher distillation, without blind backbone (w/o Backbone), and without gait phase and command adaptation in action (w/o Adaptation). X-axis is the difficulty of terrain. Results show that removing key components significantly degrades performance, particularly in traversing complex terrains such as gaps and hurdles, highlighting the effectiveness of the complete design.}
    \label{fig:Ablation}
\end{figure}

Table~\ref{tab:mae_terrain} presents the mean absolute error (MAE) results across seven terrain types, comparing our proposed method with several ablated variants, a CNN-based and ResNet-based baseline. \textit{w/o GRU} refers to replacing the GRU module with an MLP for processing temporal features. \textit{w/o Condition} denotes feeding the rough heightmap directly into the UNet decoder, instead of using the encoded depth latent as the conditional input. Our full model consistently achieves the lowest error across most terrains. To better understand the contribution of each module, we perform ablation studies by systematically removing individual components. Overall, the experiment validates the effectiveness of our proposed architecture. The integration of history-aware encoding, transformer-based cross-attention, and condition-driven decoding leads to robust generalization across terrain types, supporting accurate terrain understanding even under visual noise, occlusion, or irregular geometric layouts.

Success rate and traversing rate (the ratio of distance traveled before terminating relative to total distance) are reported in Fig.~\ref{fig:Ablation}. We report two key metrics: the traversing rate, represented by the bars, and the success rate, indicated by the lines.
The ablation results demonstrate the contribution of each component in enabling robust locomotion across diverse terrains. First, incorporating multi-teacher distillation and gait-command adaptation substantially improves performance in high-difficulty settings such as gaps and hurdles, as these mechanisms allow the policy to leverage complementary expert knowledge and dynamically adjust locomotion phase or stepping frequency according to terrain demands. Second, the backbone plays a critical role in providing a persistent forward-driving signal. In practice, we observe that robots without the backbone tend to remain stationary when encountering highly challenging terrains, as halting avoids failure penalties and can artificially increase the expected reward. While training a vision-based policy from scratch is indeed achievable through extensive reward shaping or curriculum design, recent studies have highlighted that such "reward hacking" (\textit{e.g.}, remaining in place) remains a persistent bottleneck, often necessitating complex auxiliary mechanisms. The backbone serves as an efficient architectural alternative to such laborious reward tuning. It prevents this conservative behavior by enforcing forward motion, thereby encouraging the robot to explore crossing strategies even under high risk. This design enables the emergence of successful behaviors for difficult terrains that would otherwise remain unexplored. The blind policy refers to a baseline relying solely on proprioception, whereas the depth-based policy denotes an end-to-end approach that operates directly on depth observations without explicit height map reconstruction. The depth-based policy is composed of all variants, like backbone, adaption and multi-teacher distillation.

\subsection{Real-world Reconstruction Results}

\begin{figure}[t]
    \centering
    \vspace{-2mm}
    \includegraphics[width=0.90\linewidth]{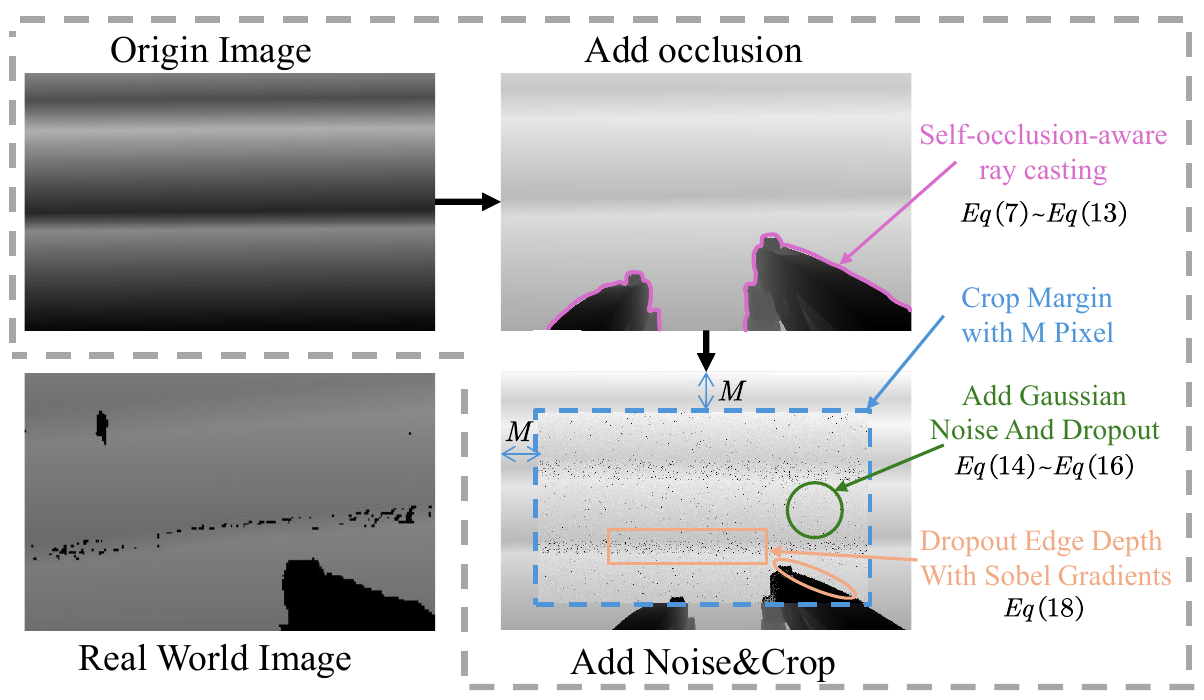}
\caption{The figure illustrates our physically grounded noise pipeline applied to synthetic depth images. From left to right: an idealized rendered image, occlusion added due to embodiment and camera angle, noise injection with boundary cropping, and a real-world depth image for comparison. The pipeline reproduces key visual artifacts observed in real sensors, including occlusion shadows, dropout, and structured noise, facilitating realistic sim-to-real transfer.}
\vspace{-2mm}
    \label{fig:depth}
\end{figure}

\begin{table}[t]
\centering
\vspace{-2mm}
\caption{Ablation of depth preprocessing components and network structure}
\scalebox{1.}{
\begin{tabular}{lc}
\toprule
\textbf{Component} & \textbf{Mean Absolute Error (cm)} \\
\midrule
Origin               & 16.07$\pm$8.38 \\
w/o Self-occlusion   & 10.07$\pm$7.48 \\
w/o Crop\&Resize        & 12.41$\pm$8.32 \\
w/o Noise Model          & 4.48$\pm$0.77 \\
CNN-based~\cite{yu2024walking} & 6.47$\pm$1.32 \\
ResNet-based~\cite{duan2024learning} & 5.31$\pm$0.82 \\
\textbf{Ours}        & \textbf{3.25$\pm$0.56} \\
\bottomrule
\end{tabular}
}
\label{tab:ablation_depth_preproc}
\end{table}
\begin{figure}[t]
    \centering
    \includegraphics[width=0.85\linewidth]{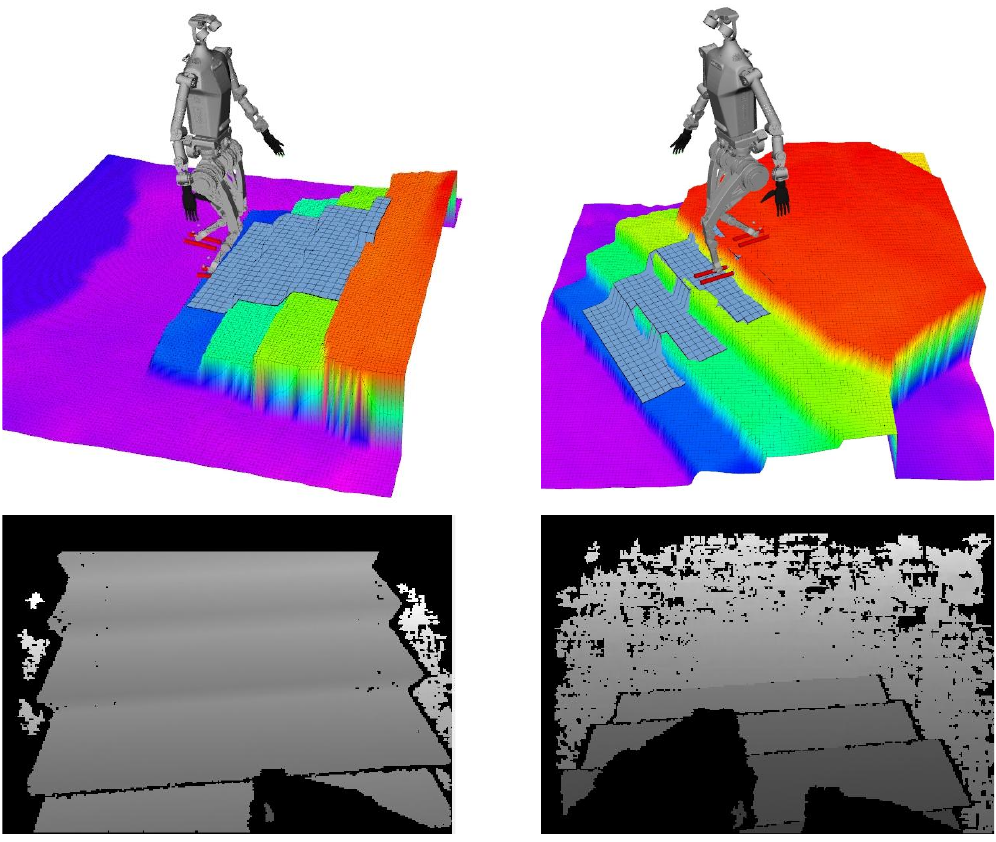}
\caption{Visual comparison of reconstructed terrain~(blue) corresponding to depth input and ground truth~(red-to-blue). The top row illustrates reconstructed terrain and ground truth built by elevation maps. The bottom row presents the corresponding raw depth images.}
    \label{fig:recon}
\end{figure}
\begin{figure}[t]
    \centering
    \vspace{-2mm}
    \includegraphics[width=0.85\linewidth]{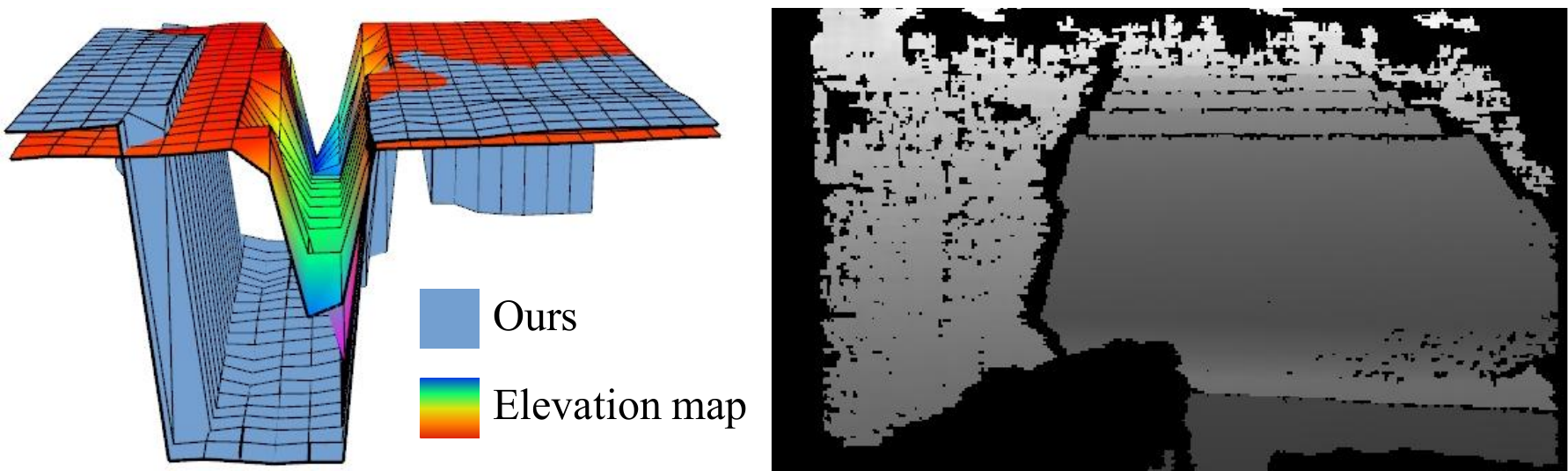}
\caption{ \textbf{Comparison between our reconstruction method and elevation map in a gap terrain scenario.} The top figure shows a 3D reconstruction of the terrain: our method (blue) successfully reconstructs the full geometry of the gap, including its bottom, while the elevation map (red–blue) fails to capture the occluded region due to missing depth information. The bottom figure shows the raw depth image, where the gap bottom is entirely occluded. Our method infers and reconstructs this missing geometry, enabling robust locomotion planning across such challenging terrains.}
    \label{fig:gap}
\end{figure}

Based on the results in Table~\ref{tab:ablation_depth_preproc}, we observe the effectiveness of incorporating geometric-aware cropping and physically grounded noise modeling in the depth pre-processing pipeline for real-world terrain reconstruction. The baseline network structures in our ablation study are all applied our depth preprocessing components.

The Origin variant processes raw depth images without any form of pre-processing, resulting in the highest reconstruction error. Removing the Crop\&Resize step moderately reduces the error, indicating that spatial normalization helps suppress peripheral distortions and improves generalization. Excluding the Noise Model, which adds both axial and lateral noise along with sigma-based and edge-aware dropout, further improves accuracy. This highlights the importance of simulating structured uncertainty patterns in the depth data. Our full method, which combines both pre-processing components, achieves the lowest MAE. These results confirm that the use of physically informed noise synthesis and structured cropping significantly enhances the robustness and accuracy of terrain reconstruction in real-world deployment settings. The total pipeline is shown in Fig.~\ref{fig:depth}

To validate the accuracy of our terrain reconstruction module, we present a visual comparison between the reconstructed elevation maps and the ground-truth terrain, as shown in Fig.~\ref{fig:recon}. The top row illustrates the reconstructed terrain (blue) alongside the ground-truth surface (colored from red to blue), both aligned in the same coordinate frame. Our reconstruction demonstrates a high degree of geometric consistency, closely matching the step structures present in the ground truth. The bottom row shows the corresponding depth images used for reconstruction, highlighting the effectiveness of our pipeline in recovering detailed and reliable terrain geometry from raw, noisy sensor observations. Such accurate reconstructions are essential for enabling robust perceptive locomotion in complex environments.
Due to the limited field of view and self-occlusion from the robot’s body, certain challenging terrains~(\eg gaps) cannot be accurately captured by the depth camera. Since the bottom of the gap remains unseen, elevation maps fail to represent such regions with sufficient fidelity. As a result, both elevation map-based methods and depth-based approaches that omit explicit reconstruction often lead to locomotion failure~\cite{chen2025learning}. We provide a visual comparison in Fig.~\ref{fig:gap} between our method and the elevation map approach, demonstrating that our method can successfully and accurately reconstruct gap-like terrain. This capability enables the locomotion policy to traverse such challenging environments reliably.
\subsection{Real-World Experiments}
\begin{table}[t]
\centering
\caption{Comparison of stumble times in stairs}
\scalebox{1.0}{
\begin{tabular}{lc}
\toprule
\textbf{Component} & \textbf{Stumble Times} \\
\midrule
Ours               & 4/10 \\
w/o end-to-end finetune   & 8/10 \\
\bottomrule
\end{tabular}
}
\label{tab:stumble}
\end{table}
\begin{figure}[t]
    \centering
    \vspace{-2mm}
    \includegraphics[width=0.85\linewidth]{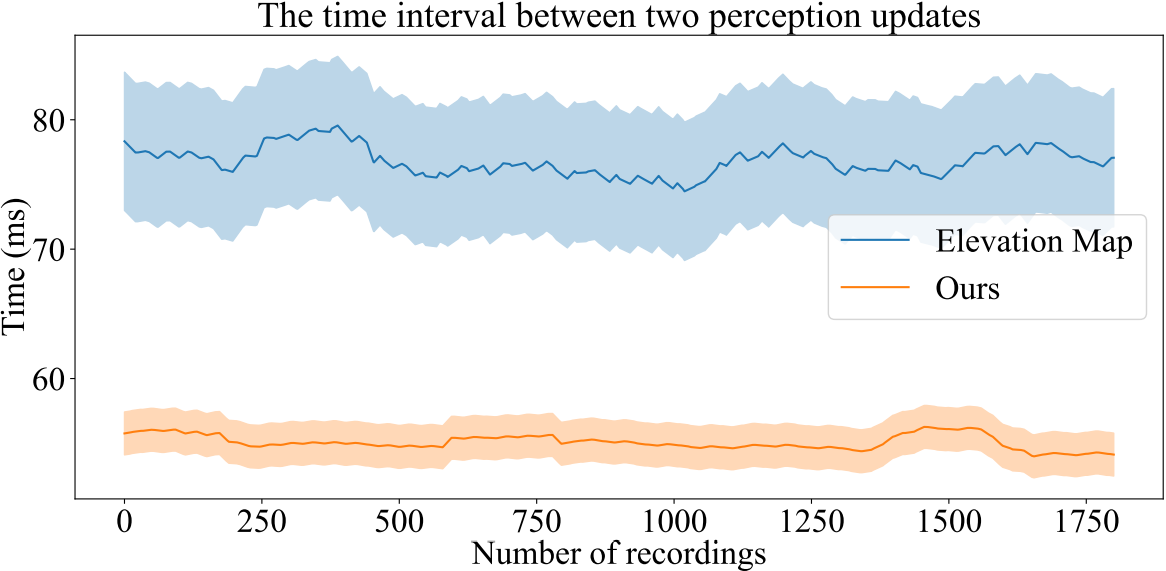}
\caption{ Comparison of the time interval between two perception updates. The proposed method achieves consistently lower and more stable update times compared with the elevation map–based approach, demonstrating improved efficiency in real-time perception.}
    \label{fig:delay}
\end{figure}
\begin{figure}[t]
    \centering
    \includegraphics[width=0.80\linewidth]{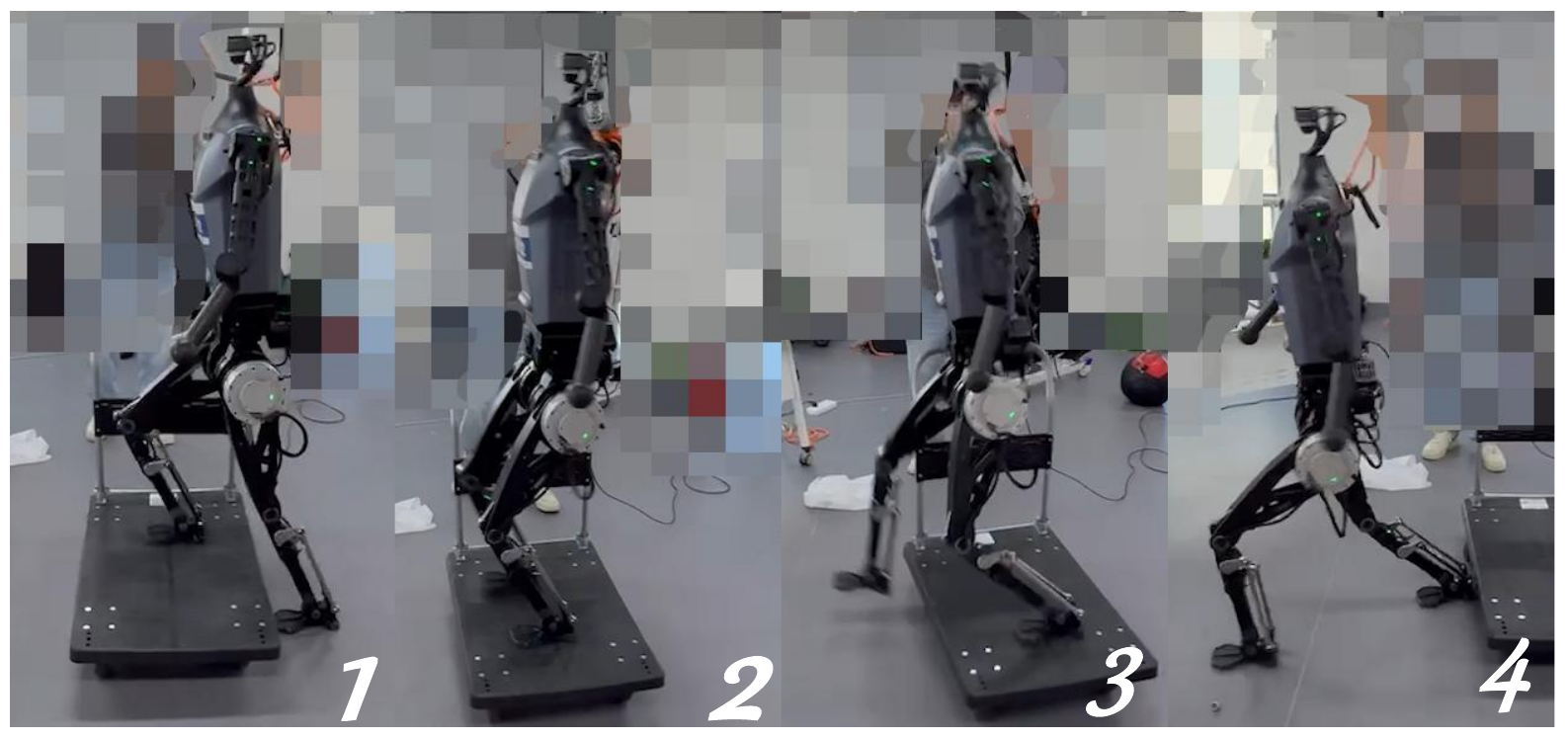}
\caption{ Sequential snapshots of the humanoid robot performing a stepping motion on a movable platform. The experiment demonstrates the robot’s capability to maintain balance and execute coordinated lower-limb movements during dynamic locomotion.}
    \label{fig:move}
\end{figure}
\begin{figure}[t]
    \centering
    \vspace{-5mm}
    \includegraphics[width=0.80\linewidth]{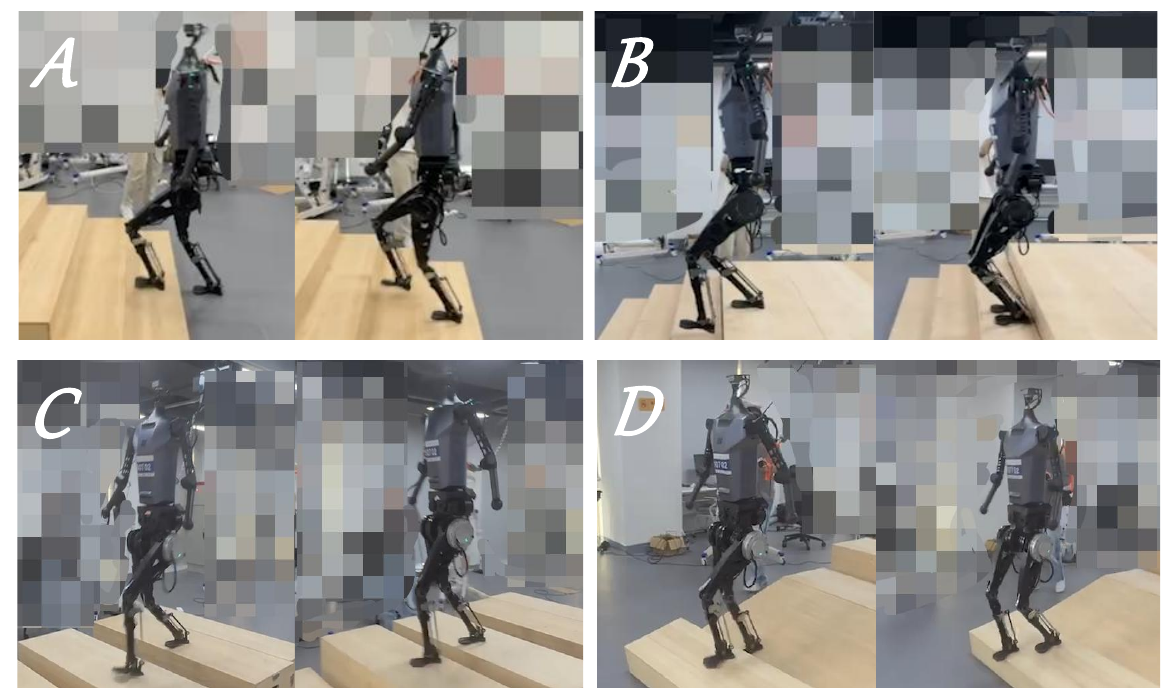}
\caption{Real-world evaluations of the humanoid robot performing adaptive locomotion across challenging terrains. 
}
\vspace{-3mm}
    \label{fig:keyframe}
\end{figure}


As shown in Fig.~\ref{fig:delay}, our method significantly reduces perception latency compared to the elevation map approach~\cite{miki2022elevation}. Operating at 30 Hz with a total delay of $\sim$20 ms, our framework outperforms the baseline, which is limited by 20 Hz LiDAR updates and an additional $\sim$30 ms mapping overhead. This lower latency and reduced variance ensure a stable perception-action loop, which is critical for minimizing the sim-to-real gap.

As shown in Fig.~\ref{fig:move}, the humanoid robot is able to perform stable stepping motions on a movable platform, despite such scenarios never being encountered during training. This experiment demonstrates that the policy can handle zero-shot deployment on previously unseen movable objects, indicating strong robustness and adaptability beyond the training distribution. As shown in Fig.~\ref{fig:keyframe}, the humanoid robot demonstrates reliable adaptability when traversing diverse real-world terrains. It successfully climbs stairs by increasing step height and maintaining balance, descends stairs with stable landing and impact mitigation, crosses consecutive gaps through step length extension and gait modulation, and directly steps down from a sloped elevated platform while preserving stability. 

Table~\ref{tab:stumble} reports the comparison of stumble frequencies in the stair traversal task. The reported data represents the cumulative number of stumbles recorded while the robot traversed a flight of 10 stair steps. Without end-to-end finetuning, the policy exhibits 8 stumbles out of 10 trials, primarily caused by the mismatch between simulated and real-world depth perception, where update delays and spatial biases result in inaccurate foot placement and frequent collisions with stair edges. In contrast, our end-to-end finetuned policy reduces stumbles to 4 out of 10 trials by explicitly adapting the reinforcement learning strategy to the frequency and latency characteristics of depth updates. This adaptation enables the robot to better anticipate perception-induced delays, thereby mitigating foot–edge collisions and significantly improving stair traversal robustness.
\vspace{-3mm}
\section{Conclusion}
We proposed a unified framework for humanoid perceptive locomotion that only utilizes a depth camera. By combining a cross-modal transformer for structured terrain reconstruction, a realistic depth images synthetic pipeline and a terrain-aware locomotion policy with a blind backbone, the system effectively mitigates perception noise, occlusion, and domain gaps while maintaining training efficiency. Real-world evaluations further validated agile and adaptive locomotion across stairs, slopes, gaps, and movable platforms, with end-to-end fine-tuning significantly reducing stumble frequencies and perception delays. These results show that integrating structured terrain reasoning with reinforcement learning provides a robust path toward reliable humanoid locomotion in unstructured environments.



\bibliographystyle{IEEEtran}
\typeout{}
\bibliography{IEEEabrv,mybibfiles}
\end{document}